\documentclass[letterpaper, 10 pt, conference]{ieeeconf}  

\IEEEoverridecommandlockouts                              

\usepackage{amsfonts}		
\usepackage{amsmath}
\usepackage[pdftex]{graphicx}
\usepackage{booktabs} 
\usepackage{subcaption}
\usepackage{multirow}
\usepackage{hyperref}
\usepackage[noadjust]{cite}

\title{\LARGE \bf
TrafficBots: Towards World Models for \\ Autonomous Driving Simulation and Motion Prediction
}

\author{Zhejun Zhang$^{1}$, Alexander Liniger$^{1}$, Dengxin Dai$^{1,2}$, Fisher Yu$^{1}$, Luc Van Gool$^{1,3}$ 
\thanks{This work is funded by Toyota Motor Europe via TRACE-Z\"urich.}
\thanks{$^{1}$Computer Vision Lab, ETH Zurich, Switzerland. {\tt\footnotesize \{zhejun.zhang, \newline alex.liniger,vangool\}@vision.ee.ethz.ch,\,i@yf.io}}%
\thanks{$^{2}$MPI for Informatics, Germany. {\tt\footnotesize ddai@mpi-inf.mpg.de}}
\thanks{$^{3}$PSI, KU Leuven, Belgium.}%
}

\begin{document}

\maketitle
\thispagestyle{empty}
\pagestyle{empty}

\begin{abstract}

Data-driven simulation has become a favorable way to train and test autonomous driving algorithms. The idea of replacing the actual environment with a learned simulator has also been explored in model-based reinforcement learning in the context of world models. In this work, we show data-driven traffic simulation can be formulated as a world model. We present TrafficBots, a multi-agent policy built upon motion prediction and end-to-end driving, and based on TrafficBots we obtain a world model tailored for the planning module of autonomous vehicles. Existing data-driven traffic simulators are lacking configurability and scalability. To generate configurable behaviors, for each agent we introduce a destination as navigational information, and a time-invariant latent personality that specifies the behavioral style. To improve the scalability, we present a new scheme of positional encoding for angles, allowing all agents to share the same vectorized context and the use of an architecture based on dot-product attention. As a result, we can simulate all traffic participants seen in dense urban scenarios. Experiments on the Waymo open motion dataset show TrafficBots can simulate realistic multi-agent behaviors and achieve good performance on the motion prediction task.

\end{abstract}


\begin{figure*}[t]
    \centering
    \includegraphics[width=0.85\textwidth]{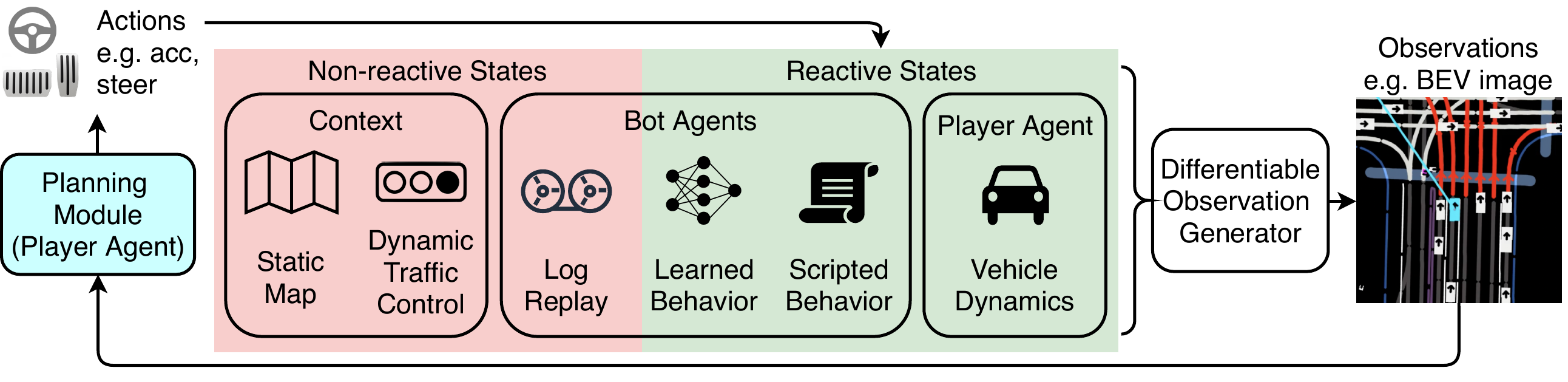}
    \vspace{-1.5ex}
    \caption{World model for AD planning modules. The simulator is fully data-driven and differentiable.}
    \label{fig:sim_overview}
    \vspace{-2ex}
\end{figure*}
\begin{figure*}[t]
    \centering
    \includegraphics[width=0.85\textwidth]{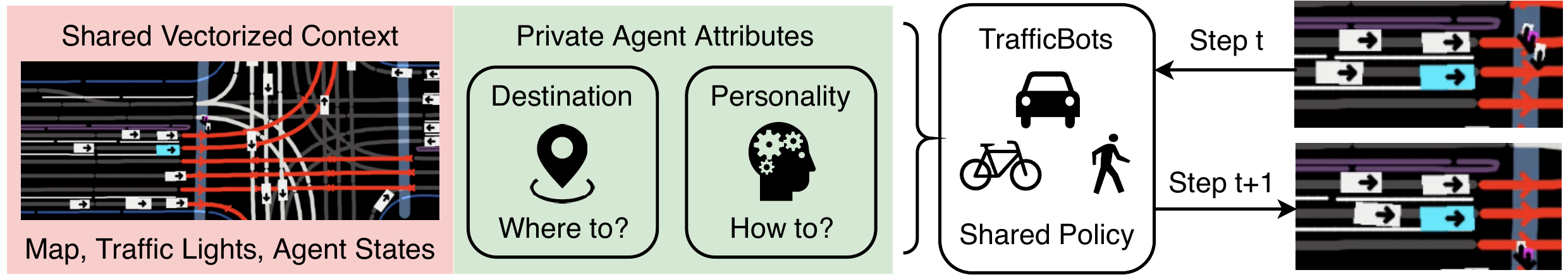}
    \vspace{-0.8ex}
    \caption{TrafficBots, a multi-agent policy that generates realistic behaviors for bot agents by learning from real-world data.}
    \label{fig:trafficbots_overview}
    \vspace{-2.5ex}
\end{figure*}

\section{Introduction}

To realize autonomous driving (AD) in the urban environment, the planning module of autonomous vehicles has to address highly interactive driving scenarios involving human drivers, pedestrians and cyclists.
Despite being a necessary step, the validation of planning algorithms on public roads is often too expensive and dangerous.
Therefore, simulations have been widely adopted and efforts have been made to develop photo-realistic driving simulators~\cite{Dosovitskiy17}.
While the full-stack simulators are popular for testing AD stacks and training visuomotor policies, they are not the best choice for developing planning algorithms because the simulated scenarios are not as sophisticated and realistic as those encountered in the real world.
Moreover, the computationally demanding rendering is redundant for AD planning modules that expect intermediate-level representations as input.

Therefore, simulators tailored for AD planning should have a different design and rely on real-world datasets. As shown in Fig.~\ref{fig:sim_overview},
the player agent, i.e., the planning module, generates a motion plan by observing some intermediate-level representations.
Then the simulator updates its internal states and generates a new observation based on the actions taken by the player agent.
The internal states of the simulation can be separated into two categories depending on whether they are reactive to the player agent.
The scenario contexts, including the map and traffic controls, are non-reactive states loaded from the datasets.
The states of the player agent are reactive and can be updated using vehicle dynamics.
Of the most importance for the simulation fidelity are the bot agents, i.e., the non-player agents.
The behaviors of bot agents fall into three categories: the non-reactive log-replay, the scripted behavior based on heuristics, and the learned behavior which is our focus.
To generate human-like behaviors for bot agents, we present TrafficBots, a multi-agent policy built upon two established research fields: multi-modal motion prediction and end-to-end (E2E) driving.

As shown in Fig.~\ref{fig:trafficbots_overview}, the TrafficBots policy is conditioned on the \emph{destination} of each agent, which approximates the output of a navigator available in the problem formulation of E2E driving~\cite{zhang2021end}.
To learn diverse behaviors from demonstrations, each TrafficBot has a \emph{personality} learned using conditional variational autoencoder (CVAE)~\cite{sohn2015Learning} following multi-modal motion prediction.
Compared to other methods, TrafficBots consume less memory, scale to more agents, and run faster than real time.
This is achieved by using a vectorized representation~\cite{gao2020Vectornet} for the context and sharing it among all bots.
A new scheme of positional encoding (PE) is introduced for angles such that the memory-efficient dot-product attention can be used to retrieve local information from the shared context that lies in the global coordinate.

Using TrafficBots and a differentiable observation generator, the simulator in Fig.~\ref{fig:sim_overview} is fully differentiable and it summarizes the player agent's past experience, hence it can be trained and used like a \emph{world model}~\cite{ha2018Recurrent}.
In this paper we focus on the TrafficBots and leave the training of player agents as future work.
We evaluate TrafficBots on both the simulation and motion prediction tasks.
We show that motion prediction can be formulated as the \emph{a priori simulation}, hence it is a legit surrogate task for the evaluation of simulation fidelity.
While prior works on traffic simulation introduce their own metrics, baselines and datasets, evaluation with motion prediction ensures an open and fair comparison.
Although our performance is not comparable to the state-of-the-art open-loop methods, TrafficBots shows the potential of solving motion prediction with a closed-loop policy.

Our contributions are summarized as follows:
We address data-driven traffic simulation using world models and we present TrafficBots, a multi-agent policy built upon motion prediction and E2E driving.
We improve the simulation configurability by introducing the navigational destination and the latent personality, as well as the scalability by introducing a new PE for angles.
Based on the public dataset and leaderboard, we propose a comprehensive and reproducible evaluation protocol for traffic simulation.
Our repository is available at \url{https://github.com/zhejz/TrafficBots}

\section{Related Work}

\textbf{World models}~\cite{ha2018Recurrent} are action-conditional dynamics models learned from observational data.
As a differentiable substitute of the actual environment, world models can be used for planning~\cite{hafner2019Learning} and policy learning~\cite{hafner2020Dream}.
In this paper, we use world models to address a new problem: traffic simulation.
We seek to obtain a world model realistic enough to replace the real world or full-stack simulators for developing AD planning algorithms.
Training world models is often formulated as a video prediction problem such that the method can generalize to all image-based environments, like Atari~\cite{hafner2021Mastering} and highway driving~\cite{henaff2019ModelPredictive}.
Although the same approach can be applied to urban driving via rasterization, this would cause unnecessary complexity because most dynamics of driving can be explicitly modeled without deep learning.
In fact, only the decision-dynamics of the bot agents that have a potential to interact with the player agent have to be learned.
To this end, we introduce the multi-agent policy TrafficBots and based on it we build a world model for AD planning.


\textbf{Motion prediction} for AD is a popular research topic.
Here we only discuss the most relevant works and refer the reader to~\cite{varadarajan2022multipath++} for a detailed review.
Our TrafficBots use a network architecture based on Transformers~\cite{vaswani2017attention} and vectorized representations~\cite{gao2020Vectornet} because they achieve top performance~\cite{ngiam2021Scene,liu2021Multimodal} while being computationally more efficient~\cite{girgis2022Latent}.
To improve the multi-agent performance, our Transformer-based architecture uses a new PE for angles.
Goal-conditioning can improve the performance of AD planning~\cite{albrecht2021interpretable, brewitt2021grit} and motion prediction~\cite{zhao2020Tnt,gu2021Densetnt,rhinehart2019Precog,deo2021Multimodal}, but it leads to causal confusions if applied to closed-loop policy.
This problem is solved by replacing the goal, which is associated with the prediction horizon, with the destination, which is time-independent and emulates a navigator.
Once conditioned on the destination, the behavior of TrafficBots agent is characterized by a time-invariant personality.
The personality is represented as the latent variable of a CVAE, which is used to address the multi-modality of motion forecasting~\cite{lee2017Desire,casas2020Implicit,tang2019Multiple,ivanovic2019trajectron}.
Unlike other works, we use a time-invariant personality, i.e., a fixed sample is used throughout the simulation horizon.
Finally, TrafficBots is related to~\cite{rhinehart2018R2p2,salzmann2020Trajectron,vazquez2022deep} in the sense that a recurrent policy is learned and combined with vehicle dynamics.
However, our method is recurrent and closed-loop, whereas motion prediction methods are open-loop.

\textbf{Data-driven simulation} can reduce the sim-to-real gap while being more efficient and scalable than manually developing a simulator.
While many works on data-driven simulation focus on the photo-realism~\cite{kim2021DriveGAN,amini2020Learning,amini2022VISTA}, we study the behavior-realism of bot agents.
Compared to the hand-crafted rules~\cite{Dosovitskiy17,SUMO2018,highway-env}, more realistic behaviors can be generated through log-replay~\cite{scheel2022Urban,caesar2021NuPlan} or learning from demonstrations~\cite{behbahani2019Learning}.
The problem of learning realistic behaviors is formulated as generative adversarial imitation learning~\cite{ho2016Generative} in~\cite{igl2022Symphony}, as behavioral cloning in~\cite{bergamini2021SimNet} and as flow prediction in~\cite{kamenev2022PredictionNet}.
Most related to our method is TrafficSim~\cite{suo2021TrafficSim}, an auto-regressive extension of the motion prediction method ILVM~\cite{casas2020Implicit}.
Compared to our method, TrafficSim is not based on world models or E2E driving, it uses rasterization and it does not factorize the uncertainty into personality and destination.
Finally, our simulation shown in Fig.~\ref{fig:sim_overview} can be considered a data-driven extension of SMARTS~\cite{zhou2020SMARTS}, and TrafficBots shown in Fig.~\ref{fig:trafficbots_overview} can be used as a sub-module to control bot agents in other simulators~\cite{zhou2020SMARTS,Dosovitskiy17,highway-env}.

\begin{figure*}[t]
    \centering
    \includegraphics[width=0.95\textwidth]{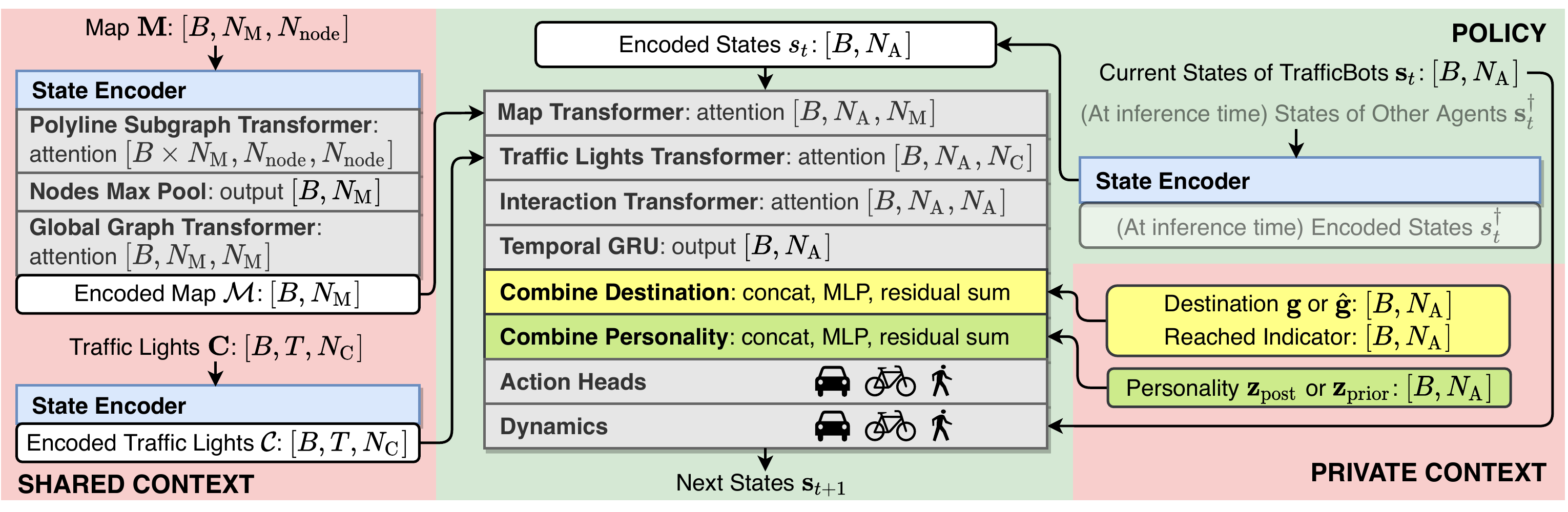}
    \vspace{-1.5ex}
    \caption{Network architecture of TrafficBots. In the brackets are the tensor shapes where $B$ is the batch size. The hidden/feature dimensions are omitted for conciseness. The shared and private contexts are encoded only once at the start of an episode.}
    \label{fig:net_and_pe}
    \vspace{-3.3ex}
\end{figure*}

\section{Problem Formulation}

We use motion prediction datasets to train a policy, which can be used for simulation if a complete episode is given, and for motion prediction if only the history is available.

\textbf{Data representation.}\;\;%
Each episode in the motion prediction dataset includes the static map $\mathbf{M}\in \mathbb{R}^{N_\text{M} \times N_\text{node} \times 4}$, traffic lights $\mathbf{C} \in \mathbb{R}^{T \times N_\text{C} \times 4}$, agent states $\hat{\mathbf{s}} \in \mathbb{R}^{T \times N_\text{A} \times 6}$ and agent attributes $\mathbf{u} \in \mathbb{R}^{N_\text{A} \times 4}$, where $N_\text{M}$ is the number of map polylines, $N_\text{node}$ is the number of nodes per polyline, $N_\text{C}$ is the number of traffic lights and $N_\text{A}$ is the number of agents.
We define $t=0$ to be the current step, $T_\text{h}$ to be the history length and $T_\text{f}$ to be the future length.
A polyline node or a traffic light is represented by $(x,y,\theta,u)$ where $x,y$ are the positions, $\theta$ is the yaw angle and $u$ is the polyline type or light state.
The ground truth (GT) state of agent $i$ at step $t$ is denoted by $\hat{\mathbf{s}}_t^i = (x,y,\theta,\dot \theta, v, a)$ where $\dot \theta$ is the yaw rate, $v$ is the speed and $a$ is the acceleration.
The time-invariant agent attribute $\mathbf{u}$ includes the agent size and type of each agent.
We use a scene-centric, vectorized representation~\cite{ngiam2021Scene} to ensure the efficiency of the simulation.

\textbf{Simulation.}\;\;%
We denote the states of TrafficBots agents as $\mathbf{s}$ and the states of other agents, including the player and other bots, as $\mathbf{s}^\dagger$.
Given a complete episode, we initialize the simulation with the history $t\in[-T_\text{h},0]$ and rollout for the future steps $t\in[1,T_\text{f}]$.
We assume all uncertainties can be explained by the GT future, thus the simulation can be formulated as predicting a single-modal next state $\mathbf{s}_{t+1}$ given $\mathbf{s}_t$ and $\mathbf{s}^\dagger_t$.
Given the GT future, the simulation has two formulations: counterfactual and a posteriori.
In \emph{counterfactual simulation} the behavior of some agents, e.g. the player agent, might deviate from the GT, i.e., $\mathbf{s}^\dagger \neq \hat{\mathbf{s}}^\dagger$. 
In this case TrafficBots should be reactive to the change and behave naturally.
In the second case, if all agents are either controlled by TrafficBots or log-replay agents, the simulation should ideally reconstruct the same episode.
In the spirit of world models, we refer to this as the \emph{a posteriori simulation}.

\textbf{Motion prediction.}\;\;%
We formulate motion prediction as the \emph{a priori simulation}, a special case of the a posteriori simulation where TrafficBots control all agents and the GT is given for $t\in[-T_\text{h},0]$.
In this case, rolling out for $t\in[1, T_\text{f}]$ is equivalent to predicting $\mathbf{s}_{1:T_{f}}^{1:N_\text{A}}$, the joint future of all agents.
Since the GT future is unavailable and multiple futures are possible given the same history, the a priori simulation is multi-modal and each rollout represents one possible way of how the scenario could evolve.
In fact, a priori simulation is equivalent to the multi-modal joint future prediction, which is a more difficult and hence less common task in comparison to the multi-modal marginal motion prediction that considers the prediction independently for each agent.

\section{TrafficBots}

As shown in Fig.~\ref{fig:net_and_pe}, the TrafficBots policy is conditioned on shared and private contexts which are encoded beforehand and explain all uncertainties, thus the rollout is deterministic.

\subsection{Policy}

The policy predicts agent states at the next step $\mathbf{s}_{t+1}$, based on the current states $\mathbf{s}_{t}$ and the contexts.
After encoding $\mathbf{s}_t$, the contexts are sequentially injected into the encoded states $s_t$.
We use Transformer encoder layers with cross-attention to update $s_t$ by attending to the encoded map $\mathcal{M}$ and the encoded traffic lights $\mathcal{C}_t$.
The interaction Transformer uses self-attention across the agent dimension to allow agents to attend to each other.
At inference time, states of non-TrafficBots agents $s_t^\dagger$ will also be processed by these Transformers such that TrafficBots can react to them.
After incorporating the map, traffic lights and states of other agents, each agent has a recurrent unit to aggregate its history because the simulation states are not Markovian.
Then the outputs are combined with the agent's individual destination and personality via concatenation and residual MLP.
Finally, the actions of each agent are predicted by the action heads and $\mathbf{s}_{t+1}$ is computed by the dynamics module based on the actions and $\mathbf{s}_t$.

\subsection{Contexts}

\textbf{State encoder.}\;\;%
Following~\cite{ngiam2021Scene}, all shared contexts, i.e., the map $\mathbf{M}$, traffic lights $\mathbf{C}$ and agent states $\mathbf{s}$, are represented in the global coordinates and incorporated via dot-product attention.
This approach is computationally more efficient than transforming the global information to the local coordinate of each agent.
However, the dot-product attention alone cannot efficiently model a global to local coordinate transform.
To remedy this issue, PE is introduced.
Without PE, VectorNet~\cite{gao2020Vectornet} has to transform all contexts to the local coordinate of each agent.
SceneTransformer~\cite{ngiam2021Scene} concatenates the PE for position with the unit vector for direction and other attributes $u$, and then feeds it to an MLP:
\begin{align}
    & s = \text{MLP}(\text{cat}(\text{PE}(x), \text{PE}(y), \cos\theta,\sin\theta, u)), \label{eq:encoder_input} \\
     \text{with } & \text{PE}_{2i}(x) = \sin(x\cdot\omega^{\frac{2i}{d_\text{emb}}}), \,
    \text{PE}_{2i+1}(x) = \cos(x\cdot\omega^{\frac{2i}{d_\text{emb}}}), \nonumber
\end{align}
where $i\in[0,\dots,d_\text{emb}/2]$, $\omega$ is the base frequency and $d_\text{emb}$ is the embedding dimension.
This state encoder can be improved by using PE also for the direction vector~\cite{mildenhall2021nerf} and adding the PE after the MLP~\cite{vaswani2017attention}. This ends up with
\begin{equation}
    \label{eq:encoder_add}
    s = \text{cat}(\text{PE}(x), \text{PE}(y), \text{PE}(\cos\theta), \text{PE}(\sin\theta)) + \text{MLP}(u).
\end{equation}
However, empirically we observe TrafficBots using this state encoder is not sensitive to directional information.
To address this issue, we propose the following state encoder
\begin{align}
    & s = \text{cat}(\text{PE}(x), \text{PE}(y), \text{AE}(\theta), \text{MLP}(u)) \label{eq:encoder_ours} \\
    & \text{with } \text{AE}_{2i}(\theta) = \sin(\theta\cdot i),\,
    \text{AE}_{2i+1}(\theta) = \cos(\theta\cdot i) \nonumber
\end{align}
where $i\in[1,\dots,d_\text{emb}/2+1]$ and AE stands for angular encoding, a special case of sinusoidal PE we introduced to encode the radian yaw $\theta$.
Compared to PE that has to use a small $\omega$ to avoid overloading the $2\pi$ period, AE can use integer frequency because it encodes an angle.
Moreover, the addition is replaced by concatenation because other states, e.g. velocity, are highly correlated to the pose encoded by PE and AE.
We use the state encoders to encode the map, traffic lights and agent states. 
Our map encoder follows~\cite{gao2020Vectornet}, except that we use Transformers for the polyline sub-graph.

\begin{figure}
\centering
\includegraphics[width=0.45\textwidth]{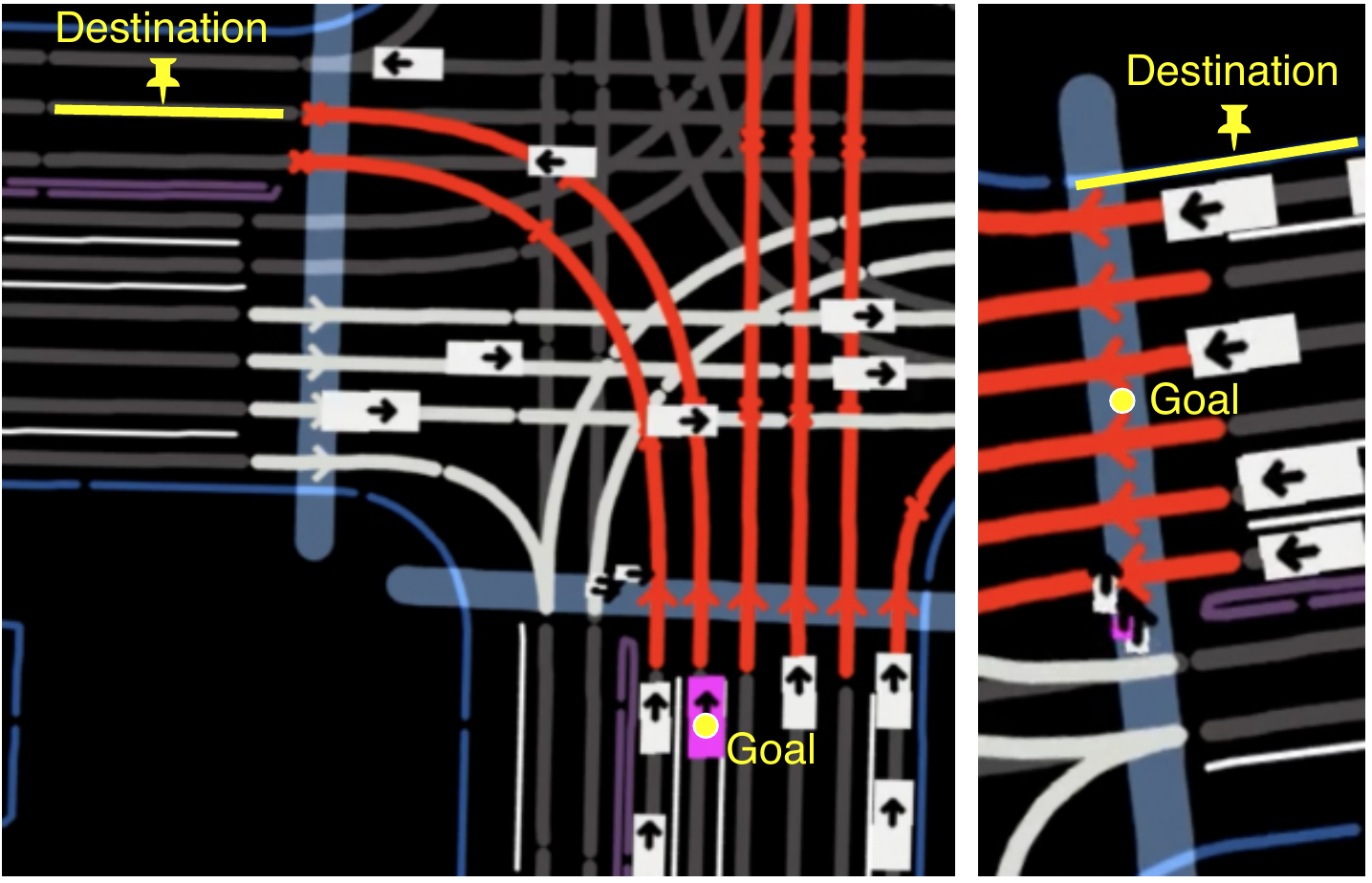}
\vspace{-1ex}
\caption{GT destination and goal of the magenta agent.}
\label{fig:destination}
\vspace{-3ex}
\end{figure}

\textbf{Destination.}\;\;%
Fig.~\ref{fig:destination} highlights the difference between our destination and the goal proposed in prior works~\cite{zhao2020Tnt,gu2021Densetnt,rhinehart2019Precog,deo2021Multimodal}.
The GT goal, which is associated to the last observed position, does not reflect an agent's intention.
In Fig.~\ref{fig:destination}, the vehicle stops because of the red light, whereas the pedestrian does not intend to stay in the middle of a crosswalk.
Although this is not a problem for open-loop motion prediction, the driving policy would learn a wrong causal relationship if conditioned on the goal.
This problem can be solved by introducing a navigator, which specifies the next goal once the current one is reached.
However, running an online navigator for every agent is computationally demanding.
For simulation with a short horizon and small maps, it is sufficient to estimate one destination for the near future, and switch to an unconditioned policy once that destination is reached.
Since the GT destination is not available in any motion prediction datasets, we approximate it with a map polyline heuristically selected by extending the recorded agent trajectory based on the map topology.
For training and simulation we use the approximated GT destination $\hat{\mathbf{g}}$, whereas for motion prediction we predict $\mathbf{g}$.
Predicting the destination is formulated as a multi-class classification task where the logit for polyline $i$ and agent $j$ is predicted by 
$\text{MLP}(\text{cat}(\mathcal{M}^i, \text{GRU}(s^j_{-T_\text{h}:0})))$, i.e., the destination of an agent depends only on the map and its own history.

\textbf{Personality.}\;\;%
In order to address the remaining uncertainties not explained by the destination and to learn diverse behaviors of different human drivers, pedestrians and cyclists, we introduce a latent personality for each agent which is learned using CVAE.
Similar ideas have been applied to world models~\cite{ha2018Recurrent,hafner2019Learning,hafner2020Dream} and motion prediction~\cite{lee2017Desire,casas2020Implicit,tang2019Multiple}.
The personality encoder has a similar architecture as the policy network in Fig.~\ref{fig:net_and_pe}.
For training and simulation, we use the posterior $\mathbf{z}_\text{post}$ which is estimated from the complete episode $t\in[-T_\text{h},T_\text{f}]$, whereas for motion prediction we use the prior $\mathbf{z}_\text{prior}$ that encodes only the history $t\in[-T_\text{h},0]$.
In contrast to TrafficSim~\cite{suo2021TrafficSim} which updates the latent at each time step to address all uncertainties, our personality is time-invariant because the behavioral style of an agent will not change in a short time horizon if the destination is determined.


\subsection{Training}
Similar to world models~\cite{ha2018Recurrent}, our training uses reparameterization gradients and back-propagation through time (BPTT).
Given a complete episode, we first encode the map $\mathbf{M}$, traffic lights $\mathbf{C}$ and GT agent states $\hat{\mathbf{s}}$.
Then we predict $\mathbf{z}_\text{post}$, $\mathbf{z}_\text{prior}$ and the destination $\mathbf{g}$.
Conditioned on the GT destination $\hat{\mathbf{g}}$ and a sample of $\mathbf{z}_\text{post}$ we rollout the policy.
For $t\in[-T_\text{h},0]$ we warm-start using teacher-forcing with GT agent states, whereas for $t\in[1,T_\text{f}]$ the rollout is auto-regressive.
All components are trained simultaneously using the weighted sum of three losses:
the reconstruction loss with smoothed L1 distance for the states $(x,y,\theta,v)$, the KL-divergence between $\mathbf{z}_\text{post}$ and $\mathbf{z}_\text{prior}$ clipped by free nats~\cite{hafner2019Learning}, and the cross-entropy loss for destination prediction.
Following~\cite{hafner2020Dream}, we stop the gradient from the action and allow only the gradient from the states during the BPTT.
We train with all agents so as to generate realistic behaviors for all traffic participants, not just for the interested ones heuristically selected by the dataset.

\begin{table}
\setlength{\tabcolsep}{3.9pt}
\centering
\caption{Results on the WOMD (marginal) leaderboard.}
\label{table:waymo_test}
\vspace{-1ex}
\begin{tabular}{lcccccc} 
\toprule
\emph{test}
& \begin{tabular}{@{}c@{}} mAP \\ $\uparrow$ \end{tabular} 
& \begin{tabular}{@{}c@{}} min \\ ADE $\downarrow$ \end{tabular} 
& \begin{tabular}{@{}c@{}} min \\ FDE $\downarrow$ \end{tabular} 
& \begin{tabular}{@{}c@{}} miss \\ rate $\downarrow$ \end{tabular} 
& \begin{tabular}{@{}c@{}} overlap \\ rate $\downarrow$ \end{tabular}  \\
\cmidrule(lr){1-1}\cmidrule(lr){2-6}
DenseTNT~\cite{gu2021Densetnt} &
$0.328$ & $1.039$ & $1.551$ & $0.157$ & $0.178$ \\
SceneTransformer~\cite{ngiam2021Scene} &
$0.279$ & $0.612$ & $1.212$ & $0.156$ & $0.147$ \\
MultiPath~\cite{chai2019Multipath} static &
$0.236$ & $0.880$ & $2.044$ & $0.345$ & $0.166$ \\
Waymo LSTM~\cite{ettinger2021Large} & $0.176$ & $1.007$ & $2.355$ & $0.375$ & $0.190$ \\
TrafficBots (a priori) & $0.212$ & $1.313$ & $3.102$ & $0.344$ & $\mathbf{0.145}$ \\
\midrule
\emph{valid} \quad TrafficBots
& $\uparrow$
& $\downarrow$
& $\downarrow$
& $\downarrow$
& $\downarrow$  \\
\cmidrule(lr){1-1}\cmidrule(lr){2-6}
a priori (\emph{K=6})
& $0.210$ & $1.291$ & $3.117$ & $0.346$ & $0.143$ \\
GT sdc future (\emph{what-if}) 
& $0.214$ & $1.281$ & $3.095$ & $0.342$ & $0.142$ \\
GT traffic light (\emph{v2x}) 
& $0.209$ & $1.288$ & $3.100$ & $0.345$ & $0.143$ \\
GT destination (\emph{v2v}) 
& $0.217$ & $1.292$ & $3.123$ & $0.345$ & $0.142$ \\
a posteriori (\emph{K=1}) 
& $0.332$ & $0.962$ & $2.034$ & $0.339$ & $0.129$ \\
\bottomrule
\end{tabular}
\vspace{-3ex}
\end{table}

\begin{table*}[t]
\setlength{\tabcolsep}{4pt}
\centering
\caption{Ablation on the WOMD validation split. All models are trained for 24K iterations (48 hours).}
\label{table:ablation}
\begin{tabular}{llcccccccccccc} 
\toprule
& & \multicolumn{6}{c}{a priori simulation K=6 (motion prediction)} & \multicolumn{6}{c}{a posteriori simulation K=1 } \\
\cmidrule(lr){3-8}\cmidrule(lr){9-14}
& & \begin{tabular}{@{}c@{}} mAP \\ $\uparrow$ \end{tabular} 
& \begin{tabular}{@{}c@{}} min \\ ADE $\downarrow$ \end{tabular} 
& \begin{tabular}{@{}c@{}} min \\ FDE $\downarrow$ \end{tabular} 
& \begin{tabular}{@{}c@{}} miss \\ rate $\downarrow$ \end{tabular} 
& \begin{tabular}{@{}c@{}} overl. \\ rate $\downarrow$ \end{tabular} 
& \begin{tabular}{@{}c@{}} NLL $\downarrow$ \\ $(\times 10^{-7})$ \end{tabular} 
& \begin{tabular}{@{}c@{}} dif. pos \\ (m) $\downarrow$ \end{tabular} 
& \begin{tabular}{@{}c@{}} dif. rot \\ (deg) $\downarrow$ \end{tabular} 
& \begin{tabular}{@{}c@{}} veh col \\ (\%) $\downarrow$ \end{tabular} 
& \begin{tabular}{@{}c@{}} run red \\ (\%) $\downarrow$ \end{tabular} 
& \begin{tabular}{@{}c@{}} passive \\ (\%) $\downarrow$ \end{tabular}
& \begin{tabular}{@{}c@{}} miss \\ rate $\downarrow$ \end{tabular}  \\
\cmidrule(lr){1-2}\cmidrule(lr){3-8}\cmidrule(lr){9-14}
Our best & TrafficBots &
$\mathbf{0.18}$ & $1.49$ & $3.66$ & $\mathbf{0.39}$ & $\mathbf{0.15}$ & $1.37$ &
$0.80$ & $2.84$ & $\mathbf{11.5}$ & $1.31$ & $19.1$ & $0.42$ \\
\cmidrule(lr){1-2}\cmidrule(lr){3-8}\cmidrule(lr){9-14}
\multirow{2}{*}{Encoder} &
Eq.~\ref{eq:encoder_input} &
$0.12$ & $1.74$ & $4.48$ & $0.48$ & $0.18$ & $1.90$ &
$0.74$ & $3.05$ & $14.7$ & $1.47$ & $19.4$ & $0.49$ \\
& Eq.~\ref{eq:encoder_add} &
$0.14$ & $1.62$ & $4.12$ & $0.46$ & $0.17$ & $1.48$ &
$0.74$ & $3.02$ & $13.8$ & $1.46$ & $19.3$ & $0.48$  \\
\cmidrule(lr){1-2}\cmidrule(lr){3-8}\cmidrule(lr){9-14}
\multirow{2}{*}{Personality} & w/o persona &
$0.06$ & $1.66$ & $4.09$ & $0.48$ & $0.15$ & $1.16$ &
$1.29$ & $3.63$ & $13.6$ & $1.50$ & $19.2$ & $0.53$ \\
& larger KL &
$0.15$ & $1.65$ & $4.19$ & $0.42$ & $0.17$ & $1.88$ &
$\mathbf{0.47}$ & $\mathbf{2.39}$ & $12.9$ & $1.56$ & $19.1$ & $0.24$ \\ 
\cmidrule(lr){1-2}\cmidrule(lr){3-8}\cmidrule(lr){9-14}
\multirow{3}{*}{Destination}
& w/o dest. &
$0.16$ & $1.53$ & $3.80$ & $0.40$ & $0.15$ & $1.44$ &
$0.74$ & $2.63$ & $11.8$ & $\mathbf{1.29}$ & $19.3$ & $0.41$ \\ 
& goal  &
$0.17$ & $\mathbf{1.47}$ & $\mathbf{3.44}$ & $0.40$ & $0.16$ & $2.02$ &
$0.78$ & $2.68$ & $12.3$ & $1.35$ & $20.2$ & $0.42$ \\ 
& goal w/o navi &
$0.14$ & $1.57$ & $3.83$ & $0.45$ & $0.17$ & $3.39$ &
$0.79$ & $2.97$ & $15.1$ & $1.40$ & $23.3$ & $0.49$ \\ 
\cmidrule(lr){1-2}\cmidrule(lr){3-8}\cmidrule(lr){9-14}
\multirow{2}{*}{World Model}
& w/o free nats &
$0.18$ & $1.52$ & $3.74$ & $0.40$ & $0.16$ & $1.39$ &
$0.86$ & $3.00$ & $12.6$ & $1.31$ & $19.1$ & $0.44$ \\ 
& w/ action grad.  &
$0.17$ & $1.51$ & $3.71$ & $0.41$ & $0.16$ & $1.39$ &
$0.90$ & $2.82$ & $12.6$ & $1.30$ & $19.1$ & $0.46$ \\ 
\cmidrule(lr){1-2}\cmidrule(lr){3-8}\cmidrule(lr){9-14}
\multirow{3}{*}{SimNet~\cite{bergamini2021SimNet}}
& BC w/o pers. \& dest. &
$0.01$ & $2.76$ & $7.77$ & $0.76$ & $0.21$ & $2.64$ &
$2.27$ & $7.37$ & $21.9$ & $1.59$ & $19.6$ & $0.76$ \\ 
& w/o pers. \& dest.  &
$0.02$ & $1.91$ & $4.95$ & $0.55$ & $0.15$ & $\mathbf{1.10}$ &
$1.34$ & $3.69$ & $13.6$ & $1.46$ & $19.2$ & $0.54$ \\ 
& BC &
$0.09$ & $3.11$ & $9.24$ & $0.73$ & $0.21$ & $3.34$ &
$2.99$ & $7.56$ & $33.4$ & $4.27$ & $19.3$ & $0.76$ \\ 
\cmidrule(lr){1-2}\cmidrule(lr){3-8}\cmidrule(lr){9-14}
\multirow{3}{*}{TrafficSim~\cite{suo2021TrafficSim}}
& w/o dynamics &
$0.14$ & $1.81$ & $4.37$ & $0.46$ & $0.17$ & $1.68$ &
$0.72$ & $55.18$ & $48.0$ & $1.73$ & $\mathbf{18.9}$ & $0.45$ \\ 
& inter. decoder  &
$0.17$ & $1.52$ & $3.73$ & $0.41$ & $0.16$ & $1.66$ &
$0.75$ & $2.85$ & $12.8$ & $1.46$ & $19.2$ & $\mathbf{0.22}$ \\ 
& resample pers. &
$0.14$ & $1.81$ & $4.74$ & $0.47$ & $0.16$ & $1.56$ &
$0.49$ & $2.45$ & $12.8$ & $1.55$ & $19.5$ & $0.29$ \\ 
\bottomrule
\end{tabular}
\vspace{-3ex}
\end{table*}

\subsection{Implementation Details}
We use a 16-dim diagonal Gaussian for the personality.
The action heads and dynamics have the same architecture but different parameters for vehicles, cyclists and pedestrians.
We use a unicycle model with constraints on maximum yaw rate and acceleration for all types of agents.
With a hidden dimension of 128 our model has less than 3M parameters.
Considering 64 agents, 1024 map polylines and a sampling time of 0.1 second, we can parallelize 16 simulations on one 2080Ti GPU while each rollout step takes around 10 ms, which is a magnitude faster than other methods~\cite{igl2022Symphony,bergamini2021SimNet,kamenev2022PredictionNet,suo2021TrafficSim}.


\section{Experiments}

\textbf{Dataset.}\;\;%
We use the Waymo Open Motion Dataset (WOMD)~\cite{ettinger2021Large} because compared to other datasets it has longer episode lengths and more diverse and complex driving scenarios, such as busy intersections with pedestrians and cyclists.
The WOMD is also one of the largest motion prediction datasets, consisting of $487$K episodes for training, $44$K for validation and $45$K for testing.
With a fixed sampling time of $0.1$ second, each episode is $9$ seconds long and contains $91$ steps: $T_\text{h}=10$ for the history, one for the current $t=0$, and $T_\text{f}=80$ future steps that shall be predicted.

\textbf{Tasks.}\;\;%
Ultimately we want to verify the fidelity of the \emph{counterfactual simulation}, such that the simulator can be used for training and testing planning modules.
However, once the scenario diverges from the factual recording, the GT trajectories can no longer be used for evaluation metrics.
To this end, different surrogate metrics have been proposed, such as traffic rule compliance~\cite{suo2021TrafficSim} and distribution of curvatures~\cite{igl2022Symphony}.
But these metrics cannot fully reflect the behavioral fidelity because they consider only vehicles and neglect pedestrians and cyclists.
Moreover, performing well on these metrics does not mean the behavior is human-like, in fact good performance can be achieved by a hand-crafted policy.
Alternatively we can verify the fidelity of the \emph{a posteriori simulation}, where the scenario should be reconstructed and the performance can be quantified by the distance to the GT trajectories.
But since the GT future is given, a model can achieve good performance by misusing the posterior latent to memorize the GT future, instead of learning the underlying human-like behavior.
In fact, the best possible performance can be simply achieved via log-replay.
We argue the \emph{a priori simulation}, i.e., motion prediction, together with the a posteriori simulation is a better evaluation setup.
For a priori simulation, the model predicts multiple futures of how an episode might evolve.
While all predictions should demonstrate natural behaviors, at least one of them should reconstruct the GT future.
Importantly, motion prediction is usually formulated as an open-loop problem.
Although TrafficBots can be used for motion prediction by formulating it as the a priori simulation, the performance will be affected by the covariant shift and compounding errors~\cite{ross2011reduction} caused by the closed-loop rollout.
Nevertheless, we show the potential of solving motion prediction with a multi-agent policy.

\textbf{Metrics.}\;\;%
For motion prediction we follow the metrics of WOMD~\cite{ettinger2021Large}, including the accuracy metrics mAP, the distance-based minADE/FDE and miss rate, and the surrogate metric overlap rate.
Inspired by~\cite{ivanovic2019trajectron}, we further examine the sampling-based negative log-likelihood (NLL) of the GT scene.
The WOMD specifies up to $8$ agents that shall be predicted and allows up to K=6 predictions.
Accordingly, we generate 6 rollouts, i.e., the joint future of all agents, by sampling the destination and the prior personality.
For a posteriori simulation, only one rollout is generated using the most likely posterior personality and the GT destination.
The simulation fidelity is evaluated using traffic rule violation rate and distance to GT trajectories.
The differences in position and rotation are averaged over all steps and agents, whereas the rates of collision, running a red light and passiveness (stop moving for no reason) are for vehicles only.

\textbf{Comparison with motion prediction methods.}\;\;
In the first half of Table~\ref{table:waymo_test} we compare TrafficBots with open-loop motion prediction methods on the Waymo (marginal) motion prediction leaderboard.
In terms of mAP we are better than the Waymo LSTM baseline~\cite{ettinger2021Large}, but worse than other methods because TrafficBots is not optimized to generate diverse predictions which is favored by the mAP metrics.
Although the miss rates are comparable, the minADE/FDE of our method are significantly higher than other methods.
This can be explained by the compounding errors caused by the auto-regressive policy rollout.
While this drawback is well-known for closed-loop methods, TrafficBots still has its advantage which is shown by the reduced overlap rate.
Compared to the open-loop methods, it is easier for a policy to learn the correct causal relationship.
The second half of Table~\ref{table:waymo_test} shows that the prediction performance can be improved given additional information.
Since the predictions are generated via rollout, we can set some of the future observations to their GT.
For example, for conditional motion prediction (\emph{what-if}) the future trajectory of the self-driving-car is given.
Furthermore, the future traffic light states and the destinations could be obtained via vehicle-to-everything (\emph{v2x}) or vehicle-to-vehicle (\emph{v2v}) communication.
Having access to all future information, the a posteriori simulation achieves the best performance with a single (\emph{K=1}) prediction.

\textbf{Ablations.}\;\;%
In Table~\ref{table:ablation} we ablate the state encoders, personality, destination and world-model training techniques on both the a priori and the a posteriori simulation.
Our state encoder Eq.~\ref{eq:encoder_ours} with AE performs overall better than \emph{Eq.~\ref{eq:encoder_input}} and \emph{Eq.~\ref{eq:encoder_add}}.
\emph{Without the personality}, the policy is unable to capture the diverse behaviors of different traffic participants.
If we allow a \emph{larger KL} divergence by downweighting the KL loss, the performance is better for a posterior simulation but worse for motion prediction.
Then we have TrafficBots \emph{w/o destination} where the latent captures all uncertainties.
In this case the model performs worse on motion prediction because the Gaussian latent suffers from mode averaging.
If the policy is conditioned on the \emph{goal}, i.e., the polyline associated with the last observed pose, then the model will learn a wrong causal relationship and the traffic rule violation rates will increase even though the minADE/FDE are smaller.
If we use the \emph{goal w/o navigator} module that drops the goal once it is reached, the policy learning will fail completely and the performances are overall inferior.
Finally, we show world-model training techniques can improve the performance of TrafficBots.
To further compare with prior works on traffic simulation, we ablate more design differences between our method and SimNet~\cite{bergamini2021SimNet}, which uses behavioral cloning (BC) without personality or destination, as well as TrafficSim~\cite{suo2021TrafficSim}.
Generally, TrafficBots performs better but there are three interesting exceptions:
Firstly, if we allow a \emph{larger KL} or \emph{resample pers.}, the posterior will memorize the GT future and the prior will fail to infer the personality.
Consequently the model performs better for a posterior simulation but worse for motion prediction, and the traffic rule violation rates are higher because the model masters the memorization rather than the driving skills.
This highlights the importance of using a time-invariant personality and the advantage of evaluating with both a priori and a posteriori simulation.
Secondly, models without personality have smaller NLL. This is reasonable because models without CVAE generate less diverse predictions, hence the NLL is smaller.
Finally, the model with an \emph{interactive decoder} following TrafficSim~\cite{suo2021TrafficSim} shows a smaller miss rate during a posteriori simulation.
This is achieved by adding the private contexts before the interaction Transformer, such that private contexts are shared among all agents.
However, this requires the personality and destination of all agents to be known before the rollout, which is infeasible if the simulation includes a player agent whose future actions are undetermined.

\begin{figure}
     \centering
     \begin{subfigure}[b]{0.45\textwidth}
         \centering
         \includegraphics[width=\textwidth]{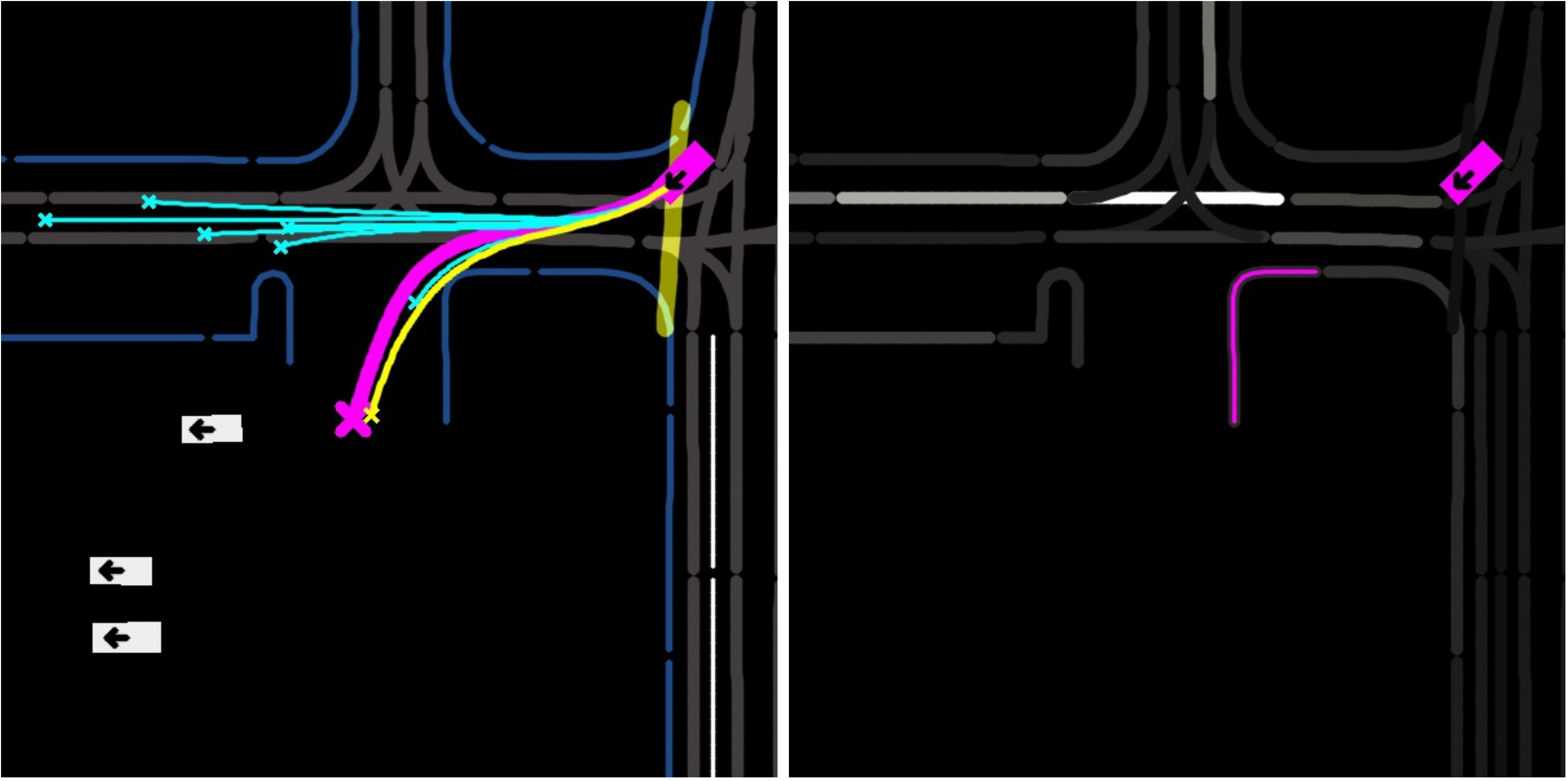}
         \vspace{-3.3ex}
         \caption{A vehicle entering the parking lots.}
         \label{fig:pred_veh}
     \end{subfigure}
     \begin{subfigure}[b]{0.45\textwidth}
         \centering
         \vspace{1ex}
         \includegraphics[width=\textwidth]{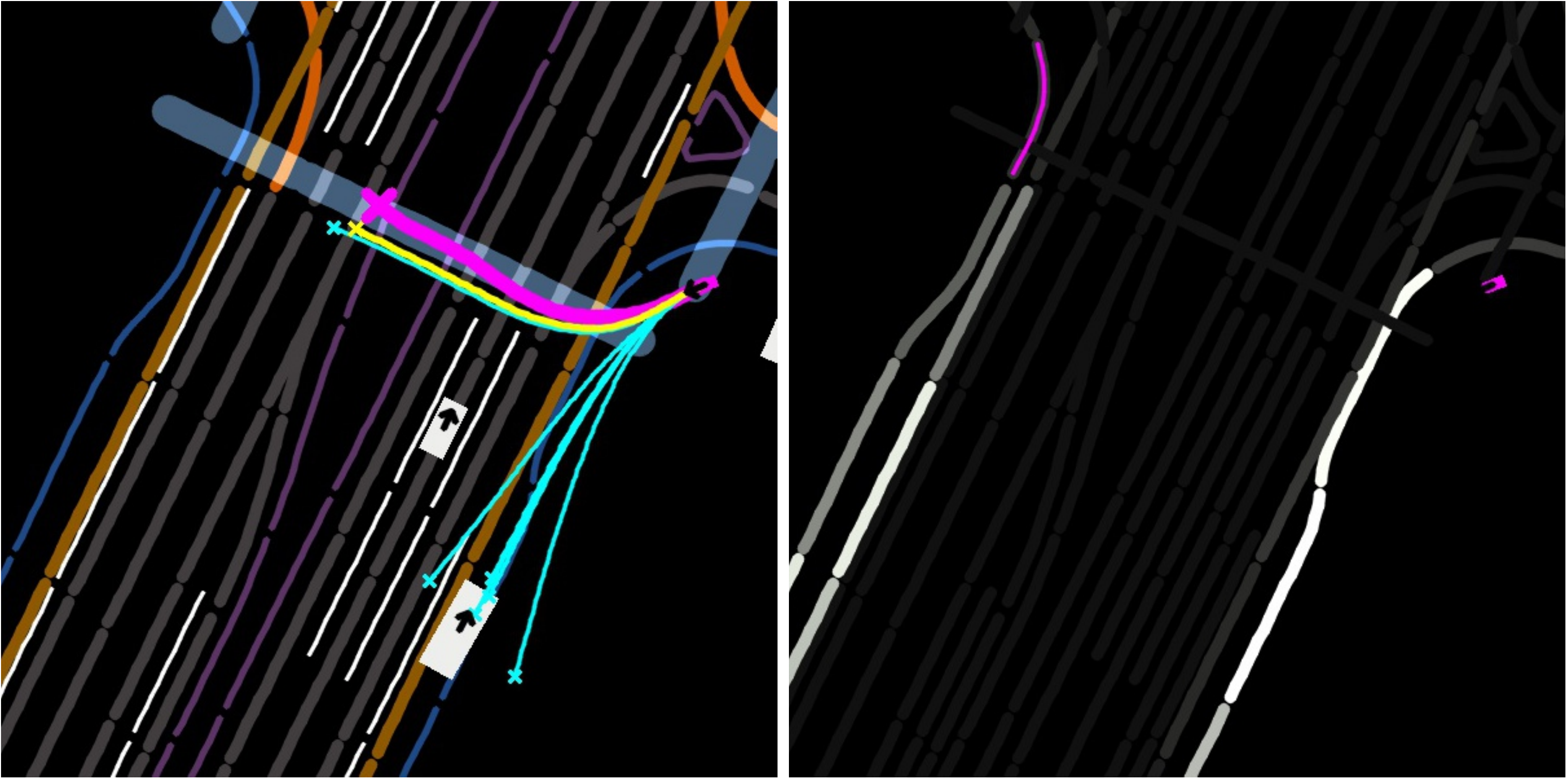}
         \vspace{-3.3ex}
         \caption{A cyclist crossing the road through the crosswalk.}
         \label{fig:pred_cyc}
     \end{subfigure}
    \caption{
    In each sub-figure, left: predicted trajectories; right: heat map of predicted destinations.
    Agent of interest and GT are in magenta. A priori predictions are in cyan.
    A posteriori simulated trajectory is in yellow.
    The brightness is proportional to the probability in the destination heat map.
    }
    \vspace{-1ex}
    \label{fig:pred}
\end{figure}

\textbf{Qualitative results.}\;\;%
Fig.~\ref{fig:pred} shows two examples of the prediction and simulation results.
In both cases, one of the a priori predictions matches the GT, whereas the a posteriori simulation reconstructs the scenario with less deviation.
With similar destinations but sampled personalities, five predictions in Fig.~\ref{fig:pred_veh} follow the lane with different speeds and lane selections.
With predicted destinations on both sides of the road, the cyclist in Fig.~\ref{fig:pred_cyc} is predicted to either cross the road or follow the road edge.

\section{Conclusions and Future Works}
This paper presented TrafficBots, a multi-agent policy learned from motion prediction datasets.
Based on the shared, vectorized context and the individual personality and destination, TrafficBots can generate realistic multi-agent behaviors in dense urban scenarios.
Besides the simulation, TrafficBots can also be used for motion prediction.
Evaluating on motion prediction tasks allows us to verify the simulation fidelity and benchmark on a public leaderboard.
Based on TrafficBots, we build a differentiable, data-driven simulation framework, which in the future can serve as a platform to develop AD planning algorithms, or as a world model to train E2E driving policies via reinforcement learning~\cite{zhang2021end} or model-based imitation learning~\cite{scheel2022Urban}.
Moreover, TrafficBots could also be integrated as a module to generate human-like behaviors for bot agents in a game or a full-stack AD simulator.
Future work will investigate better network architectures and training techniques, the downstream tasks, and combining data-driven traffic simulation with neural rendering.




\clearpage
\bibliographystyle{IEEEtran}
\bibliography{IEEEabrv,references.bib}


\appendix

\subsection{Supplementary Video}

The supplementary video for the paper is found here at~\url{https://youtu.be/2idvJOqbXeo}.
This video contains more experimental results generated by TrafficBots. The video is nicely edited and exhaustively commented. It includes two episodes, the first episode highlights a vehicle making an U-turn on a narrow street, whereas the second episode is at a busy intersection with traffic lights and a large number of traffic participants. For each episode, we first show the results of a posteriori simulation and a priori motion prediction, and then we inspect agents demonstrating the most interesting behaviors. Besides the good cases, this video also presents the bad cases where our method failed to generated realistic behavior.

\subsection{Dataset and Pre-Processing}
We use the unfiltered 9-second datasets (\emph{scenario}, not the filtered \emph{tf\_example}) from the WOMD, these are: \emph{testing, testing\_interactive, training, validation, validation\_interactive}.
The WOMD also provides a full-length training dataset \emph{training\_20s}, which includes the original 20-second-long episodes.
In contrast to the 9-second datasets which are clipped from the 20-second-long episodes, episodes in the \emph{training\_20s} do not always have a fixed length.
Although we did not use the 20-second dataset, future works can take advantage of it for simulation with a longer time horizon.
We pre-process the dataset by first filtering the map polylines:
\begin{enumerate}
  \item 
  Split the original map polylines into shorter polylines with maximum $N_\text{node}=20$ nodes one meter away from each other.
  \item 
  Remove polylines too far away from any agents.
  \item 
  Remove polylines that contain too few nodes.
  \item 
  Continue removing polylines based on the distance to agents, until the number of remaining polylines is smaller than a threshold $N_\text{M}=1024$.
\end{enumerate}

Then we filter the traffic lights which are associated with map polylines.
A traffic light will be filtered if its map polyline is removed.
Finally we filter agents as follows:
\begin{enumerate}
  \item 
  Remove agents that are tracked for too few steps.
  \item 
  Remove agents that have small displacement and large distance to any of the relevant agents marked by the WOMD or any of the map polylines. 
  These agents are mostly parking vehicles.
  \item 
  Remove vehicles that have small displacement but large yaw change, which are caused by tracking errors.
  \item 
  Continue removing irrelevant agents based on the distance to relevant agents, until the number of remaining agents is smaller than a threshold $N_\text{A}=64$.
\end{enumerate}
After the filtering, we center the episode such that the position of the self-driving-car is at $(0,0)$.
The training episodes are randomly rotated by an angle between $-\pi$ and $\pi$, whereas the validation and testing episodes are unaffected.
We smooth the agent trajectories and fill in the missing steps via temporal linear interpolation.
A pre-processed episode has $T=91$ steps and includes the following data:
\begin{enumerate}
   \item Agent states
   \begin{itemize}
      \item $agent/valid$: $[T,N_\text{A}]$, Boolean mask.
      \item $agent/pos$: $[T,N_\text{A},2]$, $x,y$ positions.
      \item $agent/vel$: $[T,N_\text{A},2]$, velocities in $x,y$ directions.
      \item $agent/spd$: $[T,N_\text{A},1]$, m/s
      \item $agent/acc$: $[T,N_\text{A},1]$, m/$\text{s}^2$
      \item $agent/yaw\_bbox$: $[T,N_\text{A},1]$, rad.
      \item $agent/yaw\_rate$: $[T,N_\text{A},1]$, rad/s.
   \end{itemize}
   \item Agent attributes
   \begin{itemize}
      \item $agent/type$: $[N_\text{A},3]$; vehicle, pedestrian, cyclist.
      \item $agent/role$: $[N_\text{A},3]$, 3 types of role; self-driving-car, agent of interest, agent to predict.
      \item $agent/size$: $[N_\text{A},3]$, length, width, height.
   \end{itemize}
   \item Map
   \begin{itemize}
      \item $map/valid$: $[N_\text{M},N_\text{node}]$, Boolean mask.
      \item $map/type$: $[N_\text{M},11]$, 11 types of polylines. They are \emph{freeway}, \emph{surface\_street}, \emph{stop\_sign}, \emph{bike\_lane}, \emph{road\_edge\_boundary}, \emph{road\_edge\_median}, \emph{solid\_single}, \emph{solid\_double}, \emph{passing\_double\_yellow}, \emph{speed\_bump} and \emph{crosswalk}.
      \item $map/pos$: $[N_\text{M},N_\text{node},2]$, $x,y$ position of nodes.
      \item $map/dir$: $[N_\text{M},N_\text{node},2]$, a 2D vector pointing to the next node.
   \end{itemize}
   \item Stop point of traffic lights
   \begin{itemize}
      \item $tl\_stop/valid$: $[T,N_\text{C}]$, Boolean mask.
      \item $tl\_stop/state$: $[T,N_\text{C}, 5]$, 5 types of states; \emph{unknown}, \emph{stop}, \emph{caution}, \emph{go} and \emph{flashing}.
      \item $tl\_stop/pos$: $[T,N_\text{C}, 2]$, position of the stop point.
      \item $tl\_stop/dir$: $[T,N_\text{C},2]$, direction of the stop point.
   \end{itemize}
\end{enumerate}

\begin{figure*}[t]
    \centering
    \includegraphics[width=0.98\textwidth]{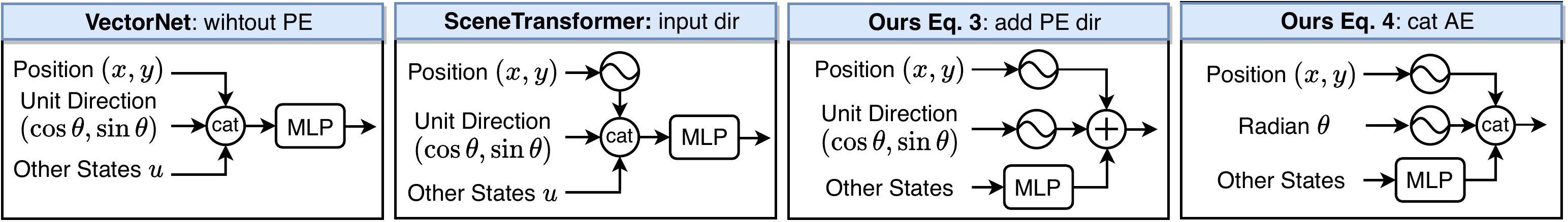}
    \caption{State encoders with different architectures.}
    \label{fig:appx_se}
    \vspace{-2.5ex}
\end{figure*}

\begin{figure}
\centering
\includegraphics[width=0.33\textwidth]{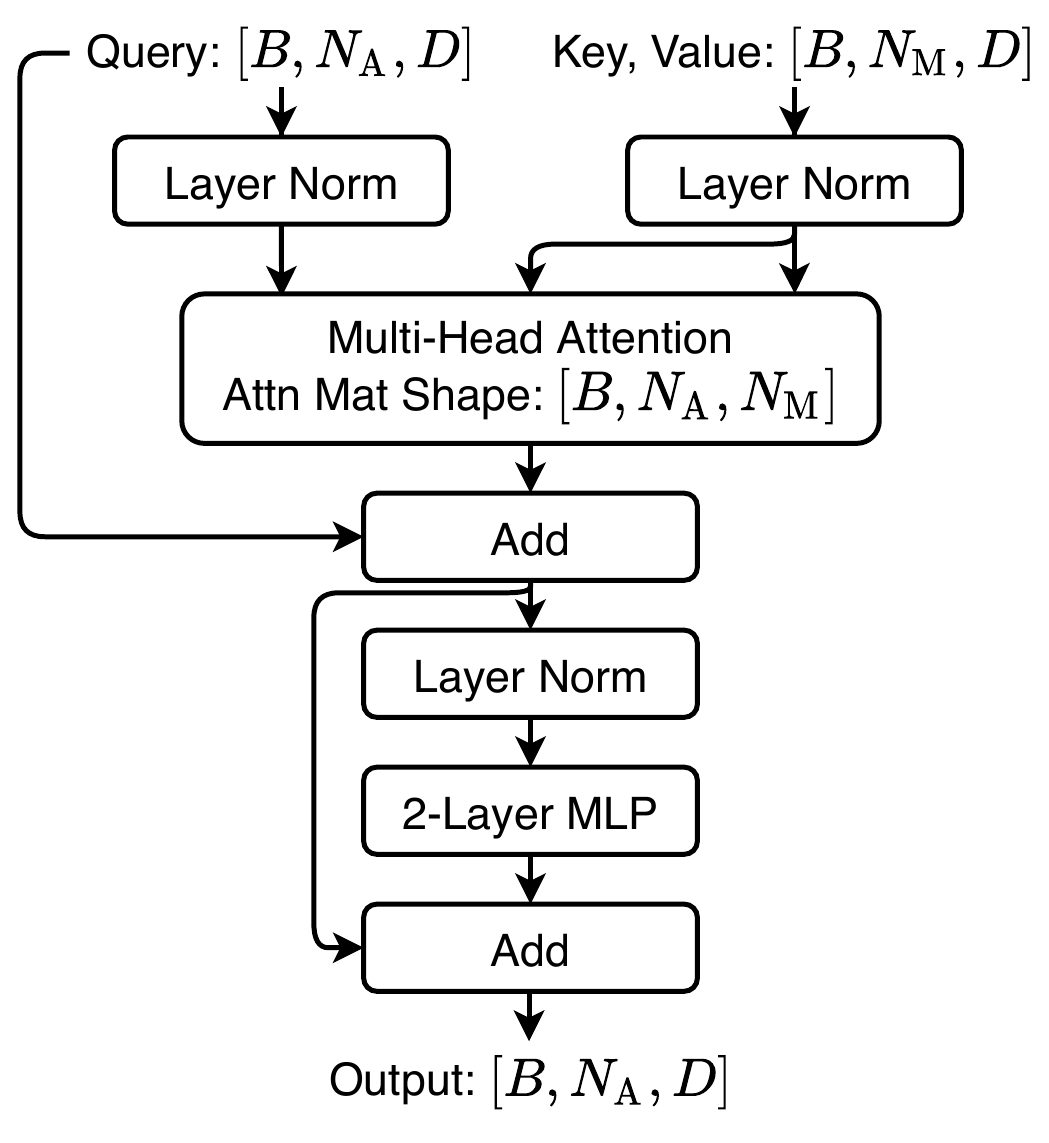}
\vspace{-2ex}
\caption{Transformer encoder layer with pre-layer-norm.}
\label{fig:appx_transformer}
\vspace{-2.5ex}
\end{figure}

\begin{figure}
\centering
\includegraphics[width=0.28\textwidth]{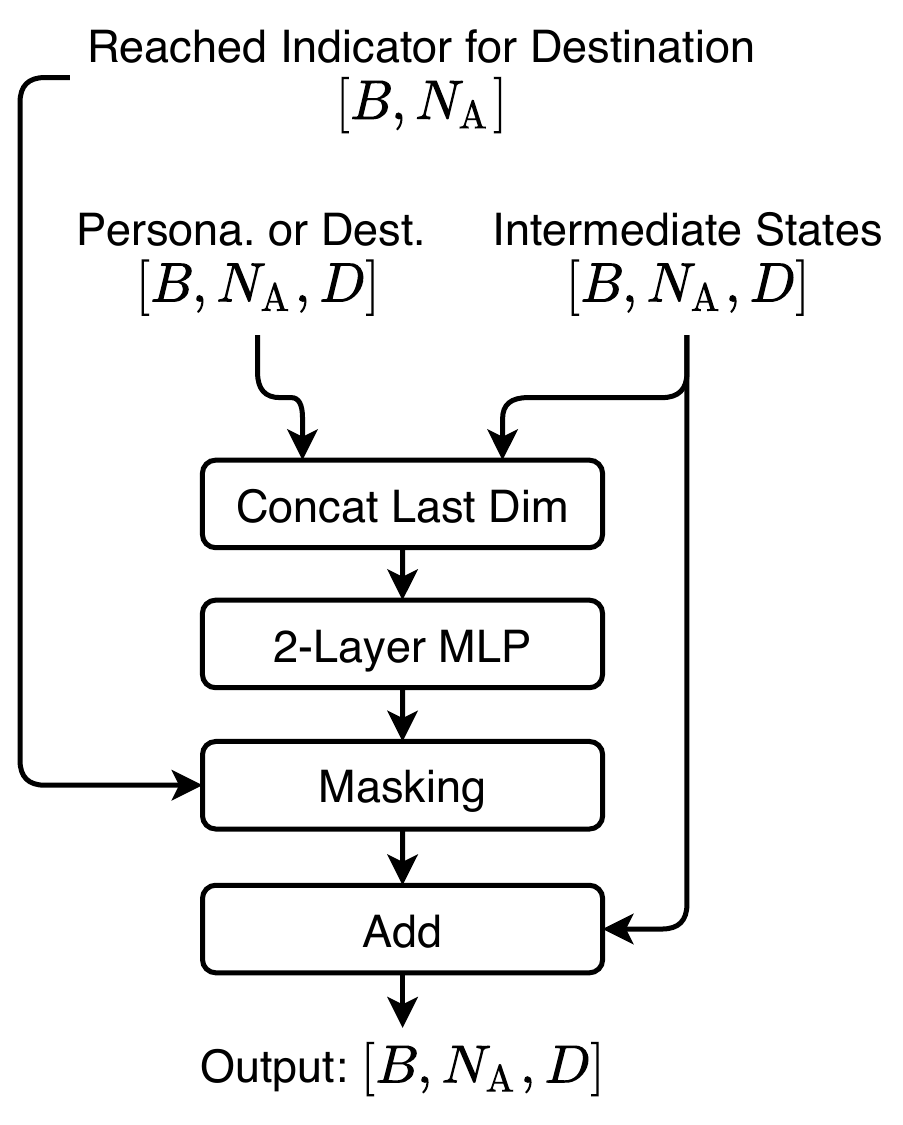}
\vspace{-1.5ex}
\caption{Combine personality/destination.}
\label{fig:appx_combine}
\vspace{-3.5ex}
\end{figure}


\subsection{Ground-Truth Destination}
The GT destinations are not available in any motion prediction datasets.
Therefore, we use the following heuristics to approximate the GT destination of an agent:
\begin{itemize}
  \item 
  If the agent is a \textbf{vehicle on a lane}, i.e. the last observed pose of the agent is close enough to a driving lane in terms of position and direction, then we find the destination by randomly selecting one of the successors of that lane base on the map topology.
  This step will be repeated multiple times.
  These agents are vehicles driving on the road.
  In this case the type of the destination is either \emph{freeway}, \emph{surface\_street} or \emph{stop\_sign}.
  \item 
  If the agent is a \textbf{vehicle not on lane}, then we extend the last observed pose with constant velocity for 5 seconds.
  After that the \emph{road\_edge\_boundary} polyline closest to that extended position will be selected.
  These agents are mostly vehicles in parking lots.
  \item 
  For \textbf{cyclists on bike lanes}, we extend the last position with constant velocity and find the closest \emph{bike\_lane}.
  \item 
  For \textbf{cyclists not on bike lanes} or \textbf{pedestrians}, we find the \emph{road\_edge\_boundary} polyline closest to the position extended using constant velocity.
\end{itemize}

For the ablation we have a model trained with \emph{goal} instead of destination.
In this case the goals are still polylines and the GT goals are still approximated using the aforementioned method, with the exception that we do not extend the last observed position using map topology or constant velocity.
The map polyline closest to the last observed position will be directly used as the \emph{goal}.
Unlike motion prediction methods~\cite{zhao2020Tnt,gu2021Densetnt} that predict an accurate goal and then simply fit a smooth trajectory towards the goal, our destination is less informative such that the motion profile is determined solely by the policy.
The destinations are pre-processed and saved as agent attributes $agent/dest: [N_\text{A}]$.
We save the indices of the corresponding map polyline, hence the value of $agent/dest$ ranges from $1$ to $N_\text{M}$.

\subsection{Detailed Network Architecture}
We use dropout probability 0.1 and ReLU activation.

\textbf{State Encoders.}\;\;
The architecture of the state encoders discussed in the main paper are visualized in Fig.~\ref{fig:appx_se}.

\textbf{Transformers.}\;\;
We use the Transformer encoder layer with cross-attention as shown in Fig.~\ref{fig:appx_transformer}.
The layer norm is inside the residual blocks~\cite{xiong2020layer}.
If the query, key and value share the same tensor, then the cross-attention boils down to self-attention which is used by the interaction Transformer.

\textbf{Combine Personality and Destination.}\;\;
As shown in Fig.~\ref{fig:appx_combine}, the personality, or the destination, is injected to the intermediate state via concatenation, MLP and residual sum.
Since the personality is always valid, the masking is unnecessary for combining personality.
In terms of destination, the masking is based on a reached indicator.
If the destination is reached, then the output of the residual block will be masked such that the intermediate states remain unchanged and the destination no longer affects the policy.

\textbf{Action Heads.}\;\;
We use a two-layer MLP to predict the acceleration and the yaw rate of each agent.
We instantiate three action heads with the same architecture; one for each type of agent.
The outputs of action heads are normalized to $[-1,1]$ via the $\tanh$ activation.

\textbf{Dynamics.}\;\;
Following MultiPath++~\cite{varadarajan2022multipath++} we use a unicycle dynamics with constraints on maximum yaw rate and acceleration for all types of agents.
For vehicles the acceleration is limited to $\pm5\,\text{m}/\text{s}$ and the yaw rate is limited to $\pm1.5\,\text{rad}/\text{s}$.
For cyclists we use $\pm6\,\text{m}/\text{s}$, $\pm3\,\text{rad}/\text{s}$ and for pedestrians $\pm7\,\text{m}/\text{s}$, $\pm7\,\text{rad}/\text{s}$.
The outputs of action heads are multiplied by the maximum allowed acceleration or yaw rate to obtain the final actions.

\begin{table*}[t]
\setlength{\tabcolsep}{8pt}
\centering
\caption{Performance on the Waymo (joint) interactive prediction leaderboard}
\label{table:waymo_int_test}
\vspace{-1.5ex}
\begin{tabular}{lcccccc} 
\toprule
\emph{test} & soft mAP $\uparrow$
& mAP $\uparrow$
& minADE $\downarrow$
& minFDE $\downarrow$
& miss rate $\downarrow$
& overlap rate $\downarrow$ \\
\cmidrule(lr){1-1}\cmidrule(lr){2-7}
DenseTNT~\cite{gu2021Densetnt} &
N/A & $0.165$ & $1.142$ & $2.490$ & $0.535$ & $0.231$ \\
SceneTransformer (J)~\cite{ngiam2021Scene} &
N/A & $0.119$ & $0.977$ & $2.189$ & $0.494$ & $0.207$ \\
Air2~\cite{wu2021air2} &
N/A & $0.096$ & $1.317$ & $2.714$ & $0.623$ & $0.247$ \\
HeatIRm4~\cite{mo2021HeatIR} &
N/A & $0.084$ & $1.420$ & $3.260$ & $0.722$ & $0.284$ \\
Waymo LSTM~\cite{ettinger2021Large} &
N/A & $0.052$ & $1.906$ & $5.028$ & $0.775$ & $0.341$ \\
TrafficBots (a priori) &
$0.113$ & $0.111$ & $1.669$ & $4.514$ & $0.681$ & $0.220$ \\
\midrule
\emph{valid} \quad TrafficBots & soft mAP $\uparrow$
& mAP $\uparrow$
& minADE $\downarrow$
& minFDE $\downarrow$
& miss rate $\downarrow$
& overlap rate $\downarrow$  \\
\cmidrule(lr){1-1}\cmidrule(lr){2-7}
a priori (\emph{K=6})
& $0.102$ & $0.100$ & $1.670$ & $4.514$ & $0.677$ & $0.221$ \\
GT sdc future (\emph{what-if}) 
& $0.110$ & $0.108$ & $1.577$ & $4.317$ & $0.651$ & $0.215$ \\
GT traffic light (\emph{v2x}) 
& $0.102$ & $0.100$ & $1.663$ & $4.485$ & $0.675$ & $0.221$ \\
GT destination (\emph{v2v}) 
& $0.106$ & $0.103$ & $1.640$ & $4.440$ & $0.668$ & $0.223$ \\
a posteriori (\emph{K=1}) 
& $0.188$ & $0.188$ & $1.085$ & $2.313$ & $0.602$ & $0.165$ \\
\bottomrule
\end{tabular}
\vspace{-4ex}
\end{table*}

\textbf{Personality Encoder.}\;\;
The inputs to the map, traffic lights and interaction Transformer of the personality encoder are reshaped differently.
For the map encoder, we flatten the agent states tensor with shape $[B,T,N_\text{A}]$ to $[B,T \times N_\text{A}]$ and use it to query the map with shape $[B,N_\text{M}]$.
This allows each agent at each time step to attend to the map independently.
For the traffic lights Transformer, the agent states tensor with shape $[B,T,N_\text{A}]$ is flattened to $[B \times T,N_\text{A}]$ and the traffic lights with shape $[B,T,N_\text{C}]$ is flattened to $[B \times T,N_\text{C}]$.
In this case, the agents states can only attend to the traffic lights from the same time step.
Similarly, inputs to the interaction Transformer are reshaped from $[B,T,N_\text{A}]$ to $[B \times T,N_\text{A}]$, such that an agent can only attend to other agents' states from the same time step.
We use two personality encoders with the same architecture to encode the posterior and the prior personality respectively.

\textbf{Latent Distribution of Personality.}\;\;
The personality encoder predicts the mean of a 16-dimensional diagonal Gaussian for each agent.
The standard deviation is a learnable parameter independent of any inputs.
We initialize the log standard deviation to $-2$ for all the 16 dimensions.
The standard deviation parameter is shared by agents from the same type (vehicle, pedestrian, cyclist).

\textbf{Predicting Destinations.}\;\;
The destination of agent $j$ depends only on the encoded map $\mathcal{M}$ and its own encoded history states $s^j_{-T_\text{h}:0}$.
Given $\mathcal{M}^i$, the hidden feature of the $i$th polyline of the encoded map $\mathcal{M}$, the logit $p^j_i$ for polyline $i$ and agent $j$ is predicted by 
\begin{align*}
    & p^j_i =\text{MLP}(\mathcal{M}^i,\text{GRU}(s^j_{-T_\text{h}:0})), \\
    \text{where } & i \in\{1,\dots,N_\text{M}\}, \, j\in\{1,\dots,N_\text{A}\}.
\end{align*}
Based on these logits, the destinations of agent $j$ are represented by a categorical distribution with $N_\text{M}$ classes and the probability is obtained via softmax.
After obtaining the polyline index $i$, the predicted destination $\mathbf{g}$ is the encoded polyline feature $\mathcal{M}^i$ indexed by $i$.
The polyline indices of the GT destinations are saved during the dataset pre-processing.

\subsection{Training Details}
We use six 2080Ti GPUs for the training with a batch size of 4 on each GPU, i.e. the total batch size is $B=24$.
Due to the large size of the WOMD training dataset, in each epoch we randomly select 15\% from the complete training and validation datasets.
We use the Adam optimizer with a learning rate of 4e-4.
The learning rate is halved every 7 epochs.
The model converges after about 30 epochs, that is almost a week.
We predict the posterior personality $\mathbf{z}_\text{post}$ using the posterior personality encoder and information from $t\in[-T_\text{h},T_\text{f}]$.
Similarly $\mathbf{z}_\text{prior}$ is predicted using the prior personality encoder and information from $t\in[-T_\text{h},0]$.
The logits of destinations are predicted using the encoded map $\mathcal{M}$ and the GT agent states $\hat s_{-T_\text{h}:0}$ from the past.
From the logits we use softmax to obtain a multi-class categorical distribution of the destination of each agent $P_\text{dest}^{1:N_\text{A}}$, which has $N_\text{M}$ classes; one for each map polyline.
During the training we rollout with the GT destination and the posterior personality $\mathbf{z}_\text{post}$.
Our training loss has the following terms:
\begin{enumerate}
  \item Reconstruction loss, which trains the model to reconstruct the GT states using the posterior personality and the GT destination. It is a weighted sum of:
  \begin{itemize}
  \item A smoothed L1 loss between the predicted $(x,y)$ positions and the GT positions.
  \item A cosine distance between the predicted yaw $\theta$ and the GT yaw $\hat \theta$, i.e. $0.5\cdot(1-\cos(\theta-\hat\theta))$.
  \item A smoothed L1 loss between the predicted velocity and the GT velocity.
  \end{itemize}
  \item The KL divergence between the posterior and the prior personality, which trains the prior to match the posterior and regularize the posterior at the same time. We use free nats~\cite{hafner2019Learning} to clip the KL divergence, i.e. if $KL(\mathbf{z}_\text{post}, \mathbf{z}_\text{prior})$ is smaller than the free nats, then the KL loss is not applied.
  We use a free nats of $0.01$.
  \item The cross entropy loss for destination classification. Since the GT destination is a single class, this loss boils down to a maximum likelihood loss, i.e. the destination distribution is trained to maximize the log-likelihood of the polyline index of the GT destination.
\end{enumerate}

\subsection{Inference Details}
We use the GT destination and the most likely posterior personality for the a posteriori simulation, hence the simulation is single modal in this case.
For a priori simulation, i.e. motion prediction, we generate multiple modes by randomly sampling the destination distribution and the prior personality of each agent.
For WOMD we generate $K=6$ predictions. 
The first mode $K_0$ is deterministic, which is generated using the most likely destination and prior personality.
We use this mode to inspect the most likely mode of the joint future prediction.
The score of each prediction, which is required by the WOMD leaderboard, is the joint probability of the destination and the personality.
We normalize the score using softmax with temperature.
The scores are computed with respect to agents, not the joint future of all agents.
For motion prediction where the future traffic light states are not available, we use the last observed (i.e. from the current step $t=0$) light states for all prediction steps.

\subsection{More Experimental Results}
In Table~\ref{table:waymo_int_test} we compare TrafficBots with other open-loop motion prediction methods on the Waymo (joint) interactive prediction leaderboard, where the joint future of exactly two agents shall be predicted and the metrics are evaluated at the scene-level, i.e. for both agents at the same time.
For a more detailed description on the task and the metrics, please refer to the publication~\cite{ettinger2021Large} or the homepage of the WOMD.
Since our method is essentially solving the joint future prediction, TrafficBots significantly outperforms the baselines on this task.
As shown in Table~\ref{table:waymo_int_test}, we achieve overall better performance than the \emph{LSTM} baseline~\cite{ettinger2021Large}.
TrafficBots also perform better than 
\emph{HeatIRm4}~\cite{mo2021HeatIR}, the winner of the 2021 WOMD challenge, and \emph{Air2}~\cite{wu2021air2}, the honorable mention of the 2021 WOMD challenge, in terms of the mAP and the overlap rate, which are the most relevant metrics used for the ranking.
Our performance is comparable to \emph{SceneTransformer (J)}, the joint version of SceneTransformer~\cite{ngiam2021Scene}.
Compared to \emph{DenseTNT}~\cite{gu2021Densetnt}, we achieve a lower overlap rate.
As discussed in the main paper, our method suffers from larger minADE/FDE and the performance can be improved given additional GT information.
These trends are also observed in Table~\ref{table:waymo_int_test}.
Although the (joint) interactive prediction is a more favorable task for our method, we do not include Table~\ref{table:waymo_int_test} in the main paper because this leaderboard is partially deprecated and hence less active, and the predictions are restricted to two agents which significantly limits its application in the real world.


\end{document}